\documentclass[conference]{IEEEtran}

\usepackage{graphicx}
\usepackage{bm}
\usepackage{amsmath}
\usepackage{amssymb}
\usepackage{esvect}
\usepackage{amsmath}
\usepackage{amssymb}
\usepackage[mathscr]{euscript}
\usepackage{csvsimple}
\usepackage{pgfplotstable}
\usepackage{nicefrac}

  \newcommand{\T}{^\mathsf{T}}
  
  \newcommand{\R}{\mathbb{R}}

  \newcommand{\N}{\mathrm{N}}

  \usepackage{bm}
  
  \newcommand{\mbf}[1]{\mathbf{#1}}
  \newcommand{\vect}[1]{\mbf{#1}}
  \newcommand{\vectb}[1]{\bm{#1}}

  \newcommand{\eg}{\textit{e.g.}}
  \newcommand{\ie}{\textit{i.e.}}
  
  \newcommand{\etc}{\textit{etc.}}
  \newcommand{\etal}{\textit{et~al.}}

  \usepackage{booktabs}

  \usepackage{tikz,pgfplots}
  \usetikzlibrary{plotmarks,shapes,arrows}
  \pgfplotsset{compat=newest} 
  \pgfkeys{/pgf/number format/.cd,1000 sep={}}

  \newlength\figureheight
  \newlength\figurewidth

  \usepackage[pagebackref=false,breaklinks=true,letterpaper=true,colorlinks,bookmarks, citecolor=black, linkcolor=black, urlcolor=black]{hyperref}

  \usetikzlibrary{external}

  \makeatletter
  \let\NAT@parse\undefined
  \makeatother

  \makeatletter
  \let\NAT@parse\undefined
  \makeatother
  \usepackage[square,numbers,sort&compress]{natbib}

  \makeatletter
  \let\MYcaption\@makecaption
  \makeatother

  \usepackage[font=footnotesize]{subcaption}

  \makeatletter
  \let\@makecaption\MYcaption
  \makeatother

\definecolor{mycolor0}{rgb}{0.2667,0.4471,0.7098}
\definecolor{mycolor1}{rgb}{0.1647,0.6706,0.3804}
\definecolor{mycolor2}{rgb}{0.8275,0.2627,0.3059}
\definecolor{mycolor3}{rgb}{0.5216,0.4392,0.7176}
\definecolor{mycolor4}{rgb}{0.8118,0.7255,0.4118}
\definecolor{mycolor5}{rgb}{0.2745,0.7176,0.8157}

\definecolor{mylcolor0}{rgb}{0.6902,0.7686,0.8863}
\definecolor{mylcolor1}{rgb}{0.5451,0.8902,0.6941}
\definecolor{mylcolor2}{rgb}{0.9412,0.7490,0.7647}
\definecolor{mylcolor3}{rgb}{0.8627,0.8392,0.9176}
\definecolor{mylcolor4}{rgb}{0.9569,0.9373,0.8667}
\definecolor{mylcolor5}{rgb}{0.7529,0.9020,0.9373}
\definecolor{mylcolor6}{rgb}{0.8750,0.8750,0.8750}

\pgfplotsset{every axis/.append style={grid style={line width=0.6pt,dotted,gray}}}

\begin{document}

\title{Iterative Path Reconstruction for Large-Scale Inertial Navigation on Smartphones}

\author{\IEEEauthorblockN{Santiago Cort\'{e}s Reina}
\IEEEauthorblockA{Aalto University\\
Espoo, Finland\\
\parbox{.2\textwidth}{\centering santiago.cortesreina@aalto.fi}}
\and
\IEEEauthorblockN{Yuxin Hou}
\IEEEauthorblockA{Aalto University\\
Espoo, Finland\\
\parbox{.2\textwidth}{\centering yuxin.hou@aalto.fi}}
\and
\IEEEauthorblockN{Juho Kannala}
\IEEEauthorblockA{Aalto University\\
Espoo, Finland\\
\parbox{.2\textwidth}{\centering juho.kannala@aalto.fi}}
\and
\IEEEauthorblockN{Arno Solin}
\IEEEauthorblockA{Aalto University\\
Espoo, Finland\\
\parbox{.2\textwidth}{\centering arno.solin@aalto.fi}}
}

\maketitle

\maketitle

\begin{abstract}
Modern smartphones have all the sensing capabilities required for accurate and robust navigation and tracking. In specific environments some data streams may be absent, less reliable, or flat out wrong. In particular, the GNSS signal can become flawed or silent inside buildings or in streets with tall buildings. In this application paper, we aim to advance the current state-of-the-art in motion estimation using inertial measurements in combination with partial GNSS data on standard smartphones. We show how iterative estimation methods help refine the positioning path estimates in retrospective use cases that can cover both fixed-interval and fixed-lag scenarios. We compare estimation results provided by global iterated Kalman filtering methods to those of a visual-inertial tracking scheme (Apple ARKit). The practical applicability is demonstrated on real-world use cases on empirical data acquired from both smartphones and tablet devices.
\end{abstract}

\section{Introduction}
\noindent
Inertial navigation systems (INS) have been studied and used for decades. The classical literature covers mainly navigation applications for aircraft, submarines and other large vehicles \cite{Jekeli:2001,Bar-Shalom+Li+Kirubarajan:2001,Titterton+Weston:2004,Britting:2010} but currently there is a large and increasing interest towards inertial navigation systems for smartphones and other light-weight consumer-grade devices (watches, tablets, drones, robots, \etc), which are often equipped with cheap and small inertial measurement units (IMUs) implemented as microelectromechanical systems (MEMS). The interest is motivated by various applications such as pedestrian navigation and wayfinding, traffic and movement analytics, robot navigation \cite{thrun_et_al:2005}, games and augmented reality.

There has been notable progress in smartphone INS by combining machine learning (\eg, learning additive and multiplicate IMU biases online \cite{Solin+Cortes+Rahtu+Kannala:2018-FUSION} or utilizing learnt priors for regressing bounds of speed \cite{cortes2018mlsp}) and additional measurements, such as automatically detected zero-velocity updates (ZUPTs), loop-closures, or manual position fixes \cite{Solin+Cortes+Rahtu+Kannala:2018-FUSION, cortes2018advio}. Thus, it has been shown that already a small number of additional infrequently occurring measurements may allow accurate motion trajectory reconstruction (by constraining the possible dynamics along the track) based on sensor streams from smartphone accelerometers and gyroscopes. The problem with the aforementioned additional measurements is that they all constrain the motion or use cases somehow. For example, zero-velocity updates do not occur if the device is handheld and constantly moving, loop-closures are hard to detect automatically and require revisiting the same locations, and manual position fixes are not automatic and require user collaboration (as in \cite{cortes2018advio}).

\begin{figure}[!t]
  \setlength{\figurewidth}{\columnwidth}
  \setlength{\figureheight}{0.6\figurewidth}
  \scriptsize
  \begin{tikzpicture}
    \node at (0,0) {\input{./fig/teaser.tex}};
    \node at (-2.5,0.2) {\includegraphics[width=1.2cm]{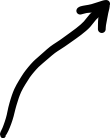}};
    \node[text width=2cm,align=center] at (-3,-1) {\bf Small-scale track from IMU};
    
    \node at (1,-1) {\includegraphics[width=1.2cm]{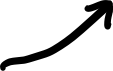}};
    \node[text width=3cm,align=center] at (0,-2) {\bf Large-scale global coordinate track from GNSS};    

  \end{tikzpicture}
  \caption{Path detail showing the general setup of our approach. The low-quality GNSS fixes (blue dots with certainty radius) tell about the global scale in world coordinates, while the accurate small-scale path details follow the IMU. Iterative re-linearization help in reconstructing the actual path shape. The thin paths show the estimate improvements over the iterations.}
  \label{fig:teaser}
\end{figure}

Hence, in order to devise a general navigation approach that would be broadly applicable, we build upon the advances in \cite{Solin+Cortes+Rahtu+Kannala:2018-FUSION} and combine their pure inertial navigation approach with automatic position measurements obtained from global navigation satellite system (GNSS) using standard smartphone hardware. That is, we revisit the GNSS aided inertial navigation problem in the smartphone context and develop a GNSS+INS solution, which is able to push the boundaries of state-of-the-art navigation systems on smartphones. In particular, we utilize iterative path reconstruction techniques to address the specific challenges that plague GNSS+INS solutions in smartphones in practical use cases: low-quality inertial sensors, gaps in GNSS reception (indoors, tunnels, shadows of tall buildings, limited availability due to automatic power saving, \etc), and high uncertainty of GNSS measurements (no RTK signal available). Thus, the focus in this work is in retrospective motion trajectory reconstruction but similar ideas could be adapted and utilized also in online use cases, and also other positioning techniques could be used instead of GNSS (\eg, Wi-Fi or Bluetooth based positioning). To some extent,  global iterative path reconstruction methods have been largely overlooked in navigation as they require storing the complete inertial sensor streams for the reconstructed path. However, even if the paths are relatively long, this does not provide a significant additional overhead or trade-offs in smartphone context since smartphone IMUs are not capable of providing data with a larger frequency than 100~Hz (iPhones) or some hundreds of Hz (Android) in any case.

Our iterative path reconstruction approach improves the accuracy compared to simply using a forward--backward extended Rauch--Tung--Striebel (RTS, see, \eg, \cite{Sarkka:2013}) smoother (as shown in Fig.~\ref{fig:teaser}). By iterative re-linearization the global scheme provides more detailed and accurate motion trajectory estimates than has been demonstrated earlier with just the use of the platform provided GNSS (see Fig.~\ref{fig:maps} for example). For example, we can precisely estimate the motion trajectory of the phone in cases, where there is hundreds of meters of handheld movement without any GNSS reception (\ie, the tri-axis accelerometer and gyroscope signals are the only measurements during the GNSS gap). As our method is a generic INS based approach, it provides 3D motion trajectories and works in any motion (pedestrian, wheeled motion, drone) and does not require steps to be detected.

The retrospective path reconstruction using low-quality and very sparse GNSS measurements is an interesting research problem which has recently re-gained momentum due mainly to interest in consumer-grade hardware applications. For example, smartphones are often used for mapping signals for various purposes, such as non-GNSS-based indoor localization \cite{Menke+Zakhor:2015,Solin+Sarkka+Kannala+Rahtu:2016}, and the accuracy that our path estimation approach provides by interpolating the trajectory from only few position measurements could be particularly useful for mapping. Also, in aerial mapping using cameras or lidars carried by a drone, our approach could provide improved accuracy, robustness and efficiency since the often used visual reconstruction methods \cite{Pix4d} are computationally heavy and their robustness may depend on the visual scene content. In addition, our approach could be used for retrospective traffic and crowd flow analysis. Further, in-store analytics and evaluation of commercial shop layouts, sports tracking, urban planning, and optimization of walkways and traffic routes could benefit from our solution.

It should be noted that also GNSS+INS methods have been studied for decades, in a similar manner as pure inertial navigation, but again the low-quality of sensors makes our case different and requires new views on solving the path reconstruction problem. For example, there are commercial devices that provide very accurate GNSS+INS solutions, such as \cite{OXTS}. However, these devices contain much better sensors than consumer devices (\ie, industrial or tactical grade IMUs and RTK assisted GNSS receivers) and they cost thousands or tens of thousands of dollars, are bulky, and consume lots of power, and hence definitely not suited for large-scale movement analysis for regular consumers. Due to large differences in  hardware, also the requirements for computational methods are quite different in these use cases. With lower quality sensors more sophisticated algorithms are needed.

In summary, the contributions of this application paper are:
\begin{itemize}
  \item A complete global iterated extended Kalman filtering approach for obtaining precise geolocalized motion trajectories (in world coordinates) for consumer-grade mobile devices despite low-quality and large gaps in GNSS reception.
  \item A comparison between our iterative path reconstruction technique and other applicable solutions (visual-inertial odometry, direct GNSS, and single forward--backward filtering) to the same estimation problem. 
  \item Experimental analysis of the capabilities and limitations of GNSS+INS approaches on smartphones.
\end{itemize}

This paper is organized as follows. In the next section we cover related approaches for pedestrian localization and movement tracking, with a focus on consumer-grade applications. Furthermore, we also cover related work in iterated filtering approaches. Sec.~\ref{sec:methods} presents the INS system dynamics and the general idea of the iterated scheme we propose to be used in the application. The main focus of the paper is in the Experiments (Sec.~\ref{sec:experiments}). Finally, we discuss the results and possible problems that may be encountered in practical applications.

\section{Related work}
\label{sec:related}
\noindent
Inertial navigation has a long history dating back to the missile guidance systems developed during the Second World War. Lately, there has been recent renewed interest  \cite{Solin+Cortes+Rahtu+Kannala:2018-FUSION,cortes2018mlsp, Trigoni, Yan+Shan+Furukawa:2018} towards the field due to the development and popularity of light-weight consumer-grade electronic devices (such as smartphones, tablets, drones, and watches), which typically include accelerometers and gyroscopes, and also GNSS receivers. However, the limited quality of consumer-grade MEMS based inertial sensors and abrupt motions of handheld, wearable and flying small devices pose challenges, which have so far prevented generic large-scale inertial navigation solutions on these devices.

The classical inertial navigation literature is extensive (see the books \cite{Jekeli:2001,Bar-Shalom+Li+Kirubarajan:2001,Titterton+Weston:2004,Britting:2010}, for example) but is mainly focused on navigation of large vehicles with relatively high quality inertial sensors. Even though the theory is solid and general, practice has shown that a lot of hand-tailoring of methods is needed to actually get working systems. Since we focus on navigation approaches using consumer-grade sensors in small mobile devices, the literature survey below concentrates on recent work in that area. 

The survey article by Harle~\cite{Harle:2013} discusses various approaches for pedestrian dead-reckoning (PDR). The main body of previous work in this category is focused on step and heading systems (SHS, \cite{Woodman:2010,Renaudin+Combettes:2014, Kang+Han:2015, Yuan+Yu+Zhang+Wang+Liu:2015, Chen+Meng+Wang+Zhang+Tian+Yang:2015}), where legged motion is assumed and inertial sensors are utilized to estimate the heading and step length of the user. Heading and step length are then used for estimating the two-dimensional walking path by accumulating the step vectors. Such systems may work reasonably well for short and medium range PDR but they typically impose constraints for device orientation. In addition, SHS may be sensitive to changing gaits and prone to false positive steps due to hand waving \cite{Harle:2013}. Some approaches, such as the work by Xiao \etal~\cite{Xiao+Wen+Markham+Trigoni:2014}, estimate the walking direction in the device frame but this is prone to errors, especially in the presence of frequent and abrupt changes in orientation. Finally, it should be noted that step and heading systems usually assume two-dimensional legged motion. Thus they do not provide complete 3D motion trajectories and do not work with wheeled motion or flying devices.

Visual-inertial odometry (VIO, \cite{Li+Kim+Mourikis:2013, Hesch+Kottas+Bowman+Roumeliotis:2014, Bloesch+Omari+Hutter+Siegwart:2015,Solin+Cortes+Rahtu+Kannala:2018-WACV}) is a technology that utilizes video cameras in addition to inertial sensors and is capable of providing full six degree of freedom motion tracking in visually distinguishable environments. Nevertheless, in contrast to GNSS+INS approaches, VIO can not provide (geo)localization but only tracking, which is prone to drifting in long term. Moreover, heavy utilization of video camera brings limitations, which make these methods impractical for wide use in consumer applications. Firstly, VIO does not work when the camera is occluded (\eg, in the pocket or bag), the field of view is filled by moving non-static objects (\eg, in a crowd) or the environment does not contain sufficient visual features (\eg, textureless surfaces indoors). Secondly, constant capturing and processing of video frames consumes a lot of energy and compromises battery longevity.

Besides SHS and VIO approaches, there are also pure inertial navigation approaches which estimate the full motion trajectory in 3D by using foot-mounted consumer-grade inertial sensors \cite{Foxlin:2005,Nilsson+Zachariah+Skog+Handel:2013}. With foot-mounted sensors the inertial navigation problem is considerably easier than in the general case since the drift can be constrained by using zero-velocity updates, which are detected on each step when the foot touches the ground and the sensor is stationary. However, automatic zero-velocity updates are not applicable for handheld or flying devices, and the approach is not suitable to large-scale consumer use since the current solutions do not work well when the movement happens without steps (\eg, in a trolley or escalator). In addition, the type of shoes and sensor placement in the foot may affect the robustness and accuracy of estimation. A prominent example in this class of approaches is the OpenShoe project \cite{Nilsson+Zachariah+Skog+Handel:2013, Nilsson+Gupta+Handel:2014}, which actually uses several pairs of accelerometers and gyroscopes to estimate the step-by-step PDR.

On the more technical side, we apply iterative filtering methods in this paper. Kalman filters and smoothers (see, \eg, \cite{Sarkka:2013} for an excellent overview of non-linear filtering) are recursive estimation schemes and thus iterative already per definition. Iterated filtering often refers to {\em local} (`inner-loop') iterations (over a single sample period). They are used together with extended Kalman filtering as a kind of fixed-point iteration to work the extended Kalman update towards a better linearization point (see, \eg, \cite{Maybeck:1982}). The resulting iterated extended Kalman filter and iterated linearized filter-smoother can provide better performance if the system non-linearities are suitable. We however, are interested in iterative re-linearization of the dynamics and passing information over the state history for extended periods. Thus we focus on so-called {\em global} (`outer-loop') schemes, which are based on iteratively re-running of the entire forward--backward pass in the filter/smoother. These methods relate directly to other iterative global linearization schemes like the so-called Laplace approximation in statistics/machine learning (see, \eg, \cite{Gelman_et_al:2013}) or Newton iteration based methods (see, \eg, \cite{Kok:2015-newton} and references therein).

In this paper, we take a general INS approach, without assuming legged or otherwise constrained motion, and compensate the limitations of low quality IMUs by fusing them with GNSS position fixes, which may be potentially sparse and infrequent containing large gaps in signal reception. As mentioned, there are relatively few general INS approaches for consumer-grade devices. We build upon the recent work \cite{Solin+Cortes+Rahtu+Kannala:2018-FUSION}, which shows relatively good path estimation results by utilizing online learning of sensor biases and manually provided loop closures or position fixes. We improve their approach in the following two ways which greatly increase the practical applicability in certain use cases: \emph{(a)} we utilize automatic GNSS based position measurements, which do not require additional manoeuvres or cooperation from the user; and \emph{(b)} we apply iterative path reconstruction methods, which provide improved accuracy in the presence of long interruptions in GNSS signal reception.

\section{Methods}
\label{sec:methods}
\noindent
The model state for our strapdown inertial navigation scheme is chosen as follows:

\begin{equation}
  \vect{x}_k = (\vect{p}_k, \vect{v}_k, \vect{q}_k, \vect{b}_k^\mathrm{a}, \vect{b}_k^{\omega}, \vect{T}_k^\mathrm{a}),
\end{equation}
where $\vect{p}_k \in \R^3$ represents the position, $\vect{v}_k \in \R^3$ the velocity, and $\vect{q}_k$ the orientation as a unit quaternion at time step $t_k$, $\vect{b}_k^\mathrm{a}$ and $\vect{b}_k^{\omega}$ are, respectively, the additive accelerometer and gyroscope bias components, and $\vect{T}_k^\mathrm{a}$ denotes the diagonal multiplicative scale error of the accelerometer.

The dynamical model (Eq.~\ref{eq:ins-model}) is based on the assumption that position is velocity once integrated, and velocity is acceleration (with the influence of gravity removed) once integrated. The orientation of the device is tracked by integrating the gyroscope measurements. The accelerometer and gyroscope readings are inserted as control signals, and their measurement noises are seen as the process noise of the system.

The dynamical model given by the mechanization equations (see, \eg, \cite{Titterton+Weston:2004,Nilsson+Zachariah+Skog+Handel:2013} for similar model formulations) is

\begin{equation}\label{eq:ins-model}
  \begin{pmatrix}
    \vect{p}_k \\[3pt] \vect{v}_k \\[3pt] \vect{q}_k  \\[3pt] \vect{b}_k^\mathrm{a} \\[3pt]  \vect{b}_k^{\omega} \\[3pt] \vect{T}_k^\mathrm{a}
  \end{pmatrix}
  =
  \begin{pmatrix}
    \vect{p}_{k-1} + \vect{v}_{k-1}\Delta t_k \\[3pt]
    \vect{v}_{k-1} + [\vect{q}_k \otimes (\tilde{\vect{a}}_k + \vectb{\varepsilon}^\mathrm{a}_k) \otimes \vect{q}_k^\star - \vect{g}] \Delta t_k \\[3pt]
     \vect{q}_{k-1} \otimes \mathscr{Q} \{ (\tilde{\vectb{\omega}}_k + \vectb{\varepsilon}^\omega_k) \Delta t_k \}  \\[3pt]
      \vect{b}_{k-1}^\mathrm{a} \\[3pt]
        \vect{b}_{k-1}^{\omega} \\[3pt]
         \vect{T}_{k-1}^\mathrm{a}
  \end{pmatrix},
\end{equation}
where the time step length is given by $\Delta t_k = t_{k} - t_{k-1}$ (note that we {\em do not} assume equidistant sampling times), the accelerometer input is denoted by $\tilde{\vect{a}}_k$ and the gyroscope input by $\tilde{\vectb{\omega}}_k$. Gravity $\vect{g}$ is a constant vector. The symbol $\otimes$ denotes quaternion product, and the rotation update is given by the function $\mathscr{Q} \{\vectb{\omega}\}$ from $\R^3$ to $\R^4$ which returns a unit quaternion (see \cite{Titterton+Weston:2004} for details). Note the abuse of notation, where vectors in $\R^3$ are operated along side quaternions in $\R^4$, in the quaternion product, vectors in $\R^3$ are assumed to be quaternions with no real component.

The process noises associated with the inputs are modelled as i.i.d.\ Gaussian noise $\vectb{\varepsilon}^\mathrm{a}_k \sim \N(\vectb{0},\vectb{\Sigma}^\mathrm{a} \Delta t_k)$ and $\vectb{\varepsilon}^\omega_k \sim \N(\vectb{0},\vectb{\Sigma}^\omega \Delta t_k)$. The link between the observed IMU readings $\vect{a}_k$ and $\vectb{\omega}_k$ and the calibrated readings are given by:
\begin{equation}
\begin{split}
  \tilde{\vect{a}}_k &= \vect{T}_k^\mathrm{a} \, \vect{a}_k - \vect{b}^\mathrm{a}_k, \\
  \tilde{\vectb{\omega}}_k &= \vectb{\omega}_k - \vect{b}^\omega_k,
\end{split}
\end{equation}
where the multiplicative bias $\vect{T}_k^\mathrm{a}$ and the additive biases $\vect{b}^\mathrm{a}_k$ and $\vect{b}^\omega_k$ are parts of the state and have constant dynamics in Eq.~\eqref{eq:ins-model}. This means that the biases can be seen as unknown constants that need to be estimated. However, typically even these parameters are given a small process noise in order too account for slow crawl. However, without loss of generality, we leave out this process noise due to our interest in sequences with time-spans of at maximum some tens of minutes.

\subsection{Inference by extended Kalman filtering}
\noindent
The Kalman filter performs a probabilistic estimation of a dynamic state given noisy measurements. It is an optimal estimator given two key constrains, the process and measurement noises are multivariate Gaussian distributions and the dynamics and measurement functions are linear. In the case of non-linear dynamics, a good approximation is to linearize the function at the given point and perform the estimation on the linearized version, this is known as the Extended Kalman filter (EKF, see, \eg, \cite{Bar-Shalom+Li+Kirubarajan:2001,Sarkka:2013}). The EKF approximates the state distributions with Gaussians, $p(\vect{x}_k \mid \vect{y}_{1:k}) \simeq \N(\vect{x}_k \mid \vect{m}_{k}, \vect{P}_{k})$ through first-order linearizations.

The dynamics are incorporated into the {\em prediction step}:
\begin{equation} \label{eq:prediction_step}
\begin{split}
  \vect{m}_{k \mid k-1} &= \vect{f}_k(\vect{m}_{k-1 \mid k-1}, \vectb{0}), \\
  \vect{P}_{k \mid k-1} &= \vect{F}_\vect{x}(\vect{m}_{k-1 \mid k-1}) \, \vect{P}_{k-1 \mid k-1} \, \vect{F}_\vect{x}\T(\vect{m}_{k-1 \mid k-1}) + \\ &\qquad
   \vect{F}_{\vectb{\varepsilon}}(\vect{m}_{k-1 \mid k-1}) \, \vect{Q}_k \, \vect{F}_{\vectb{\varepsilon}}\T(\vect{m}_{k-1 \mid k-1}),
\end{split}
\end{equation}
where the dynamic model is evaluated with the outcome from the previous step and zero noise, and $\vect{F}_\vect{x}(\cdot)$ denotes the Jacobian matrix of $\vect{f}_k(\cdot, \cdot)$ with respect to $\vect{x}$ and $\vect{F}_{\vectb{\varepsilon}}(\cdot)$ with respect to the process noise $\vectb{\varepsilon}$. The process noise covariance is set up as $\vect{Q}_k = \mathrm{blkdiag}(\vectb{\Sigma}^\mathrm{a} \Delta t_k, \vectb{\Sigma}^\omega \Delta t_k)$.

Measurement data consists of various observations of the system state at different points in time (we follow \cite{Solin+Cortes+Rahtu+Kannala:2018-FUSION} for constraining the momentary speed). These are then combined with the model in the {\em update step}:
\begin{equation}
\begin{split} \label{eq:update}
  \vect{v}_k &= \vect{y}_k - \vect{h}_k(\vect{m}_{k \mid k-1}), \\
  \vect{S}_k &= \vect{H}_\vect{x}(\vect{m}_{k \mid k-1}) \, \vect{P}_{k \mid k-1} \, \vect{H}_\vect{x}\T(\vect{m}_{k \mid k-1}) + \vect{R}_k, \\
  \vect{K}_k &= \vect{P}_{k \mid k-1} \, \vect{H}_\vect{x}\T(\vect{m}_{k \mid k-1}) \,  \vect{S}_k^{-1}, \\
  \vect{m}_{k \mid k} &= \vect{m}_{k \mid k-1} + \vect{K}_k \, \vect{v}_k, \\
  \vect{P}_{k \mid k} &= [\vect{I}-\vect{K}_k \, \vect{H}_\vect{x}(\vect{m}_{k \mid k-1})] \, 
\vect{P}_{k \mid k-1} \, [\vect{I}-\vect{K}_k \, \vect{H}_\vect{x}(\vect{m}_{k \mid k-1})]\T \\ & \quad + \vect{K}_k \, \vect{R}_k \, \vect{K}_k\T,
\end{split}
\end{equation}
where $\vect{H}_\vect{x}(\cdot)$ denotes the Jacobian of the measurement model $ \vect{h}_k(\cdot)$ with respect to the state variables $\vect{x}$. The slightly unorthodox form of the last line is known as the Joseph's formula, which both numerically stabilizes updating the covariance and preserves symmetry.

\subsection{Extended Rauch--Tung--Striebel smoother}
\noindent
As we are also interested in the {\it a~posteriori} estimates of position, velocity and orientation for already passed states, we use a fixed-interval extended Kalman smoother of Rauch--Tung--Striebel (RTS) kind. If we denote the last available time step in the data as $t_n$, this means that the approximation gives the following conditional state estimates $p(\vect{x}_k \mid \vect{y}_{1:n}) \simeq \N(\vect{x}_k \mid \vect{m}_{k \mid n}, \vect{P}_{k \mid n})$ conditioned on all the observed data in the range $t \in [t_0, t_n]$.

The extended Rauch--Tung--Striebel {\em smoothing pass} can be written using the filtering outcome as a preliminary step. The backward iteration is started from the last filtering estimate (see, \eg, \cite{Sarkka:2013} for detailed presentation):
\begin{equation}
\begin{split}
  \vect{G}_k &= \vect{P}_{k \mid k} \, \vect{F}_{\vect{x}}\T(\vect{m}_{k \mid k}) \, \vect{P}_{k+1 \mid k}^{-1}, \\
  \vect{m}_{k \mid n} &= \vect{m}_{k \mid k} + \vect{G}_k \, [\vect{m}_{k+1 \mid n} - \vect{m}_{k+1 \mid k}], \\
  \vect{P}_{k \mid n} &= \vect{P}_{k \mid k} + \vect{G}_k \, [\vect{P}_{k+1 \mid n} - \vect{P}_{k+1 \mid k}] \, \vect{G}_k\T.
\end{split}
\end{equation}

Running the smoothing pass requires the filter means and covariances to be stored for all steps.

\subsection{A global iterated extended Kalman filter}
\label{sec:giekf}
\noindent
The idea with global iterated filtering passes is to improve the track by passing over information over the track history. Once the state estimation is performed for every time step, the desire is to improve the result by reducing the linearization error. In order to do this the mean of the initial state $\vect{m}_0$ is replaced with the result of the RTS smoother $\vect{m}_{0 \mid n}$ from the previous pass, but the covariance matrix $\vect{P}_0$ is left unchanged.

The global iterated extended Kalman filter (GIEKF) scheme is given as follows:
\begin{enumerate}
  \item {\em Initialization}: Choose a prior for the initial state $p(\vect{x}_0) \simeq \mathrm{N}(\vect{m}_0, \vect{P}_0)$.
  \item {\em First iteration $(i=0)$}: Run the EKF forward pass and the associated backward pass.
  \item {\em Subsequent iterations $(i)$}: Modify the initial state mean to agree with the previous smoother pass, $\vect{m}_0 \leftarrow \vect{m}_{0 \mid n}^{(i-1)}$, and re-run the filter--smoother passes.
  \item {\em Iterate step 3 until convergence}.
\end{enumerate}

Global iterated schemes are useful when a subset of the state is time-invariant (fixed or nearly fixed dynamics for some of the state variables) and the dynamics are sensitive to linearization errors. In this case the bias parameters are invariant and the integration of rotation rates into a unit quaternion are highly non-linear.

\begin{figure}[!t]
  \scriptsize
  \begin{subfigure}[b]{0.3\columnwidth}
    \centering
    \setlength{\figurewidth}{.6\textwidth}
    \setlength{\figureheight}{3.3055\figurewidth}  
    \input{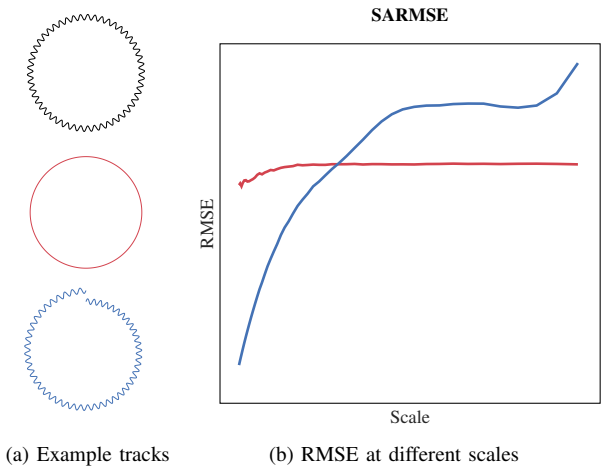}
    \vspace*{1em}
    \caption{Example tracks}
    \label{fig:toy}
  \end{subfigure}
  \begin{subfigure}[b]{0.6\columnwidth}
    \centering
    \setlength{\figurewidth}{\textwidth}
    \setlength{\figureheight}{.9\textwidth}  
%
%
\definecolor{mycolor2}{rgb}{0.2667,0.4471,0.7098}
\definecolor{mycolor1}{rgb}{0.8275,0.2627,0.3059}
\begin{tikzpicture}

\begin{axis}[%
width=0.951\figurewidth,
height=\figureheight,
at={(0\figurewidth,0\figureheight)},
scale only axis,
unbounded coords=jump,
xmin=0,
xmax=20,
xtick={\empty},
xlabel style={font=\color{white!15!black}},
xlabel={Scale},
ymin=0.1,
ymax=0.7,
ytick={\empty},
ylabel style={font=\color{white!15!black}},
ylabel={RMSE},
axis background/.style={fill=white},
title style={font=\bfseries},
title={SARMSE}
]
\addplot [color=mycolor1, line width=1.0pt, forget plot]
  table[row sep=crcr]{%
1	0.464160707022091\\
1.06	0.466906131821638\\
1.13	0.462183139723427\\
1.2	0.468004662473482\\
1.28	0.471966924450257\\
1.36	0.472339113387666\\
1.44	0.470066162891945\\
1.53	0.470507394244001\\
1.63	0.47241708997062\\
1.73	0.474093631162771\\
1.84	0.476827825316143\\
1.96	0.481287270086328\\
2.08	0.48369860271614\\
2.21	0.482117182496829\\
2.35	0.485012811917162\\
2.5	0.486913649625994\\
2.66	0.489328083956511\\
2.83	0.488663134525533\\
3.01	0.490916296241307\\
3.2	0.492947120095696\\
3.4	0.494218082417759\\
3.61	0.494980997438663\\
3.84	0.49580826640789\\
4.08	0.498061133566844\\
4.34	0.497213230839713\\
4.61	0.49769581842879\\
4.9	0.498379625249199\\
5.21	0.497945434881672\\
5.54	0.497919079312459\\
5.89	0.49901035667361\\
6.26	0.498893941542228\\
6.66	0.499001539819677\\
7.08	0.499442782515807\\
7.52	0.498520724516768\\
8	0.499141977980985\\
8.5	0.499096621956559\\
9.04	0.498696820422962\\
9.61	0.498775147591673\\
10.22	0.498574500155494\\
10.86	0.49929263171467\\
11.55	0.499461437032228\\
12.27	0.499956888826727\\
13.05	0.499478890745289\\
13.87	0.499698758436609\\
14.75	0.499262341597969\\
15.68	0.499729270837465\\
16.66	0.499878649087741\\
17.72	0.499471108128668\\
18.83	0.498737172828086\\
20.02	nan\\
};
\addplot [color=mycolor2, line width=1.0pt, forget plot]
  table[row sep=crcr]{%
1	0.163658179743033\\
1.06	0.172034219300428\\
1.13	0.181379642827795\\
1.2	0.190760333130347\\
1.28	0.201096576664729\\
1.36	0.211084360940402\\
1.44	0.220737734122291\\
1.53	0.23132428979226\\
1.63	0.242788790597548\\
1.73	0.253864736104489\\
1.84	0.265721770800765\\
1.96	0.278276365123012\\
2.08	0.290363782911493\\
2.21	0.301685935203825\\
2.35	0.314883787644226\\
2.5	0.32854735412259\\
2.66	0.340905363582486\\
2.83	0.353368293618935\\
3.01	0.365908315265066\\
3.2	0.380667291908696\\
3.4	0.393321485445166\\
3.61	0.40354486722971\\
3.84	0.416604183691155\\
4.08	0.429274919789206\\
4.34	0.439817672392538\\
4.61	0.449825158797851\\
4.9	0.46201941199612\\
5.21	0.470046537741057\\
5.54	0.480311725546279\\
5.89	0.490934933888657\\
6.26	0.50086384457713\\
6.66	0.512682988580045\\
7.08	0.526091415194538\\
7.52	0.539951893168181\\
8	0.555932420068505\\
8.5	0.570390057789911\\
9.04	0.582363645272201\\
9.61	0.590486264865839\\
10.22	0.594754364542794\\
10.86	0.596830689427564\\
11.55	0.597142782558966\\
12.27	0.599387491926081\\
13.05	0.600225046218907\\
13.87	0.600099777384937\\
14.75	0.595390062294386\\
15.68	0.593355233214737\\
16.66	0.596958077363873\\
17.72	0.61726588059132\\
18.83	0.667772995533368\\
20.02	nan\\
};
\end{axis}
\end{tikzpicture}%
    \caption{RMSE at different scales}
    \label{fig:toy-sarmse}    
  \end{subfigure}
  \caption{Example {\em Scaled Aligned RMSE} (SARMSE). In (a), the top track is the ground-truth and the two reconstructions of it are shown in red and blue. In (b), the corresponding SARMSE errors are shown for both tracks.}
  \label{fig:toy-and-error}
\end{figure}

\begin{figure*}[!t]
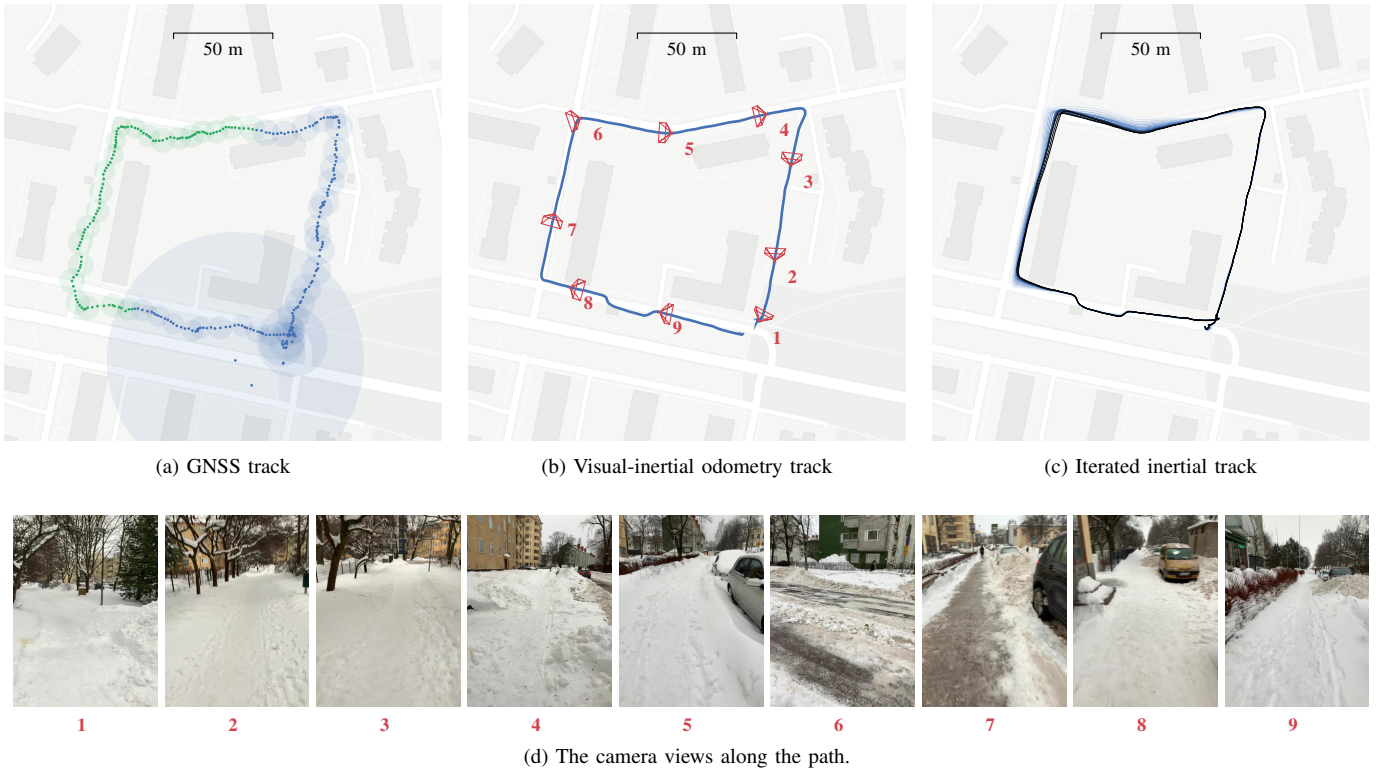

  \begin{subfigure}[b]{.32\textwidth}
    \centering
    \setlength{\figurewidth}{\textwidth}
    \setlength{\figureheight}{\textwidth}  
    \input{./fig/map-gnss.tex}
    \caption{GNSS track}
    \label{fig:map-gnss}
  \end{subfigure}
  \hfill
  \begin{subfigure}[b]{.32\textwidth}
    \centering
    \setlength{\figurewidth}{\textwidth}
    \setlength{\figureheight}{\textwidth}  
    \input{./fig/map-arkit.tex}
    \caption{Visual-inertial odometry track}
    \label{fig:map-arkit}    
  \end{subfigure}
  \hfill
  \begin{subfigure}[b]{.32\textwidth}
    \centering
    \setlength{\figurewidth}{\textwidth}
    \setlength{\figureheight}{\textwidth}  
    \input{./fig/map-ours.tex}
    \caption{Iterated inertial track}
    \label{fig:map-ours}  
  \end{subfigure}
  \\[1em] 
  \begin{subfigure}[b]{\textwidth}
    \centering
    \setlength{\figurewidth}{0.111\textwidth}
    \newcommand{\figg}[1]{\includegraphics[width=.95\figurewidth]{./fig/frame-0#1}}
    \begin{tikzpicture}
      \foreach \i in {1,...,9}
        \node[text width=.9\figurewidth,align=center,text centered,text depth = 0cm] at ({\figurewidth*\i},0) {\figg{\i}};

      \foreach \i in {1,...,9}
        \node[text width=5mm,inner sep=0,align=center,text centered,text depth = 0cm] at ({\figurewidth*\i},-.75\figurewidth) {\textcolor{mycolor2}{\scriptsize\bf \i}};
  
    \end{tikzpicture}
    \caption{The camera views along the path.}
    \label{fig:frames}
  \end{subfigure}  
  \caption{One example path captured by an Apple iPad. (a)~GNSS/platform location positions with uncertainty radius. The samples in \textcolor{mycolor1}{green} were removed for the gap experiment. (b)~The visual-inertial odometry (Apple ARKit) track that was captured for reference/validation. The ARKit fuses information from the IMU and device camera. The path has been manually aligned to the starting point and orientation. (c)~Our iterative solution of the gap experiment, where we fuse the iPad IMU readings with the \textcolor{mycolor0}{blue} GNSS locations in (a). Note that the lines straighten along the roads and the corners are square. (d)~Example frames along the path showing the test environment. Associated camera poses shown in (b). (Best viewed zoomed in.)}
  \label{fig:maps}
\end{figure*}

\subsection{Quantitative evaluation metrics for odometry tracks}
\label{sec:metric}
\noindent
Quantitative evaluation of odometry tracks is not straightforward. For example, linear interpolation of GNSS points would not compare to visual inertial odometry as the tracking for an augmented reality headset but it would be much better in the case of vehicular navigation. GNSS points are accurate but not precise, the opposite is true for visual-inertial odometry systems. In order to show these differences we consider a {\em Scaled Aligned RMSE} (SARMSE) measure, which can quantify both small-scale and large-scale error in odometry tracks. This error measure builds on the relative error measure presented in \cite{zhang+Scaramuzza:2018}.

Given a test and a ground-truth track sampled at the same points in time, the tracks are sampled at corresponding intervals (in time) of different length. Then, the segments are rigidly aligned to minimize the squared error between them. The remaining error is then averaged across the sampled segments of that given length. The result is a set of errors for a given set of the scales. Errors at low scales pertain to low accuracy while errors at large scales pertain to relative precision (with respect to the starting point).

The error for timescale $t_i$ is defined as 
\begin{equation}
e_i=\frac{1}{N}\sum_{s \in S }\| \vect{p}^\mathrm{gt}_{s}-\vect{p}^\mathrm{test}_{s}\|,
\end{equation}
where $\vect{p}^\mathrm{test}_{s:s+t_i}$ is rigidly transformed to mimnimize the RMSE with respect to $\vect{p}^\mathrm{gt}_{s:s+t_i}$ and $S$ is the set (of size $N$) of segments of length $t_i$ in the track. 

This is best explained with an example: Fig.~\ref{fig:toy} shows an example pair of tracks and the corresponding SARMSE plot. The red track is a `good' estimate of the black track but lacks the fine detail. The blue track on the other hand is also a `good' estimate of the black track but drifts away as the track moves. The SARMSE in Fig.~\ref{fig:toy-sarmse} reflects these characteristics, as the scale increases the blue error grows while the red error remains stable. The SARMSE plot tells that the red track is accurate at a large scale, and the blue at a small scale. Optimally, of course, both small-scale and large-scale error would be small.

\section{Experiments}
\label{sec:experiments}
\noindent
This paper is written application-first, and thus we have put much interest in applying the methods to empirical data captured from actual consumer devices. The focus is on Apple iPhones and iPads, mostly due to their uniform hardware, but the results should directly generalize to mid and high-range Android devices as well.

The experiments are split into two parts. The first part is a more example-driven (or even qualitative) demonstration example for one particular track. The second part then tries to capture more general use cases and variability in data by running the methods on a standard benchmark data set with 23 different sequences of both indoor and outdoor tracks.

\subsection{Example sequence in urban winter}
\noindent
We start by describing our demo sequence study, a city street capture. In this sequence, we walked around a non-square city-block in central Helsinki holding an Apple iPad Pro (11-inch, late-2018 model). Along the path there are buildings roughly 5--10 floors high. The ground was covered in snow and the side walk was completely closed for parts of the path. The user had the device pointing forward and roughly head high (for the reference visual-inertial capture, which also requires use of camera). At the beginning of the capture the user rested the smartphone against a rigid object to ensure no movement during a few frames. This is important since it allows for online calibration of the sensor biases (these are temperature sensitive so an offline version in a different environment would not work). The whole capture is roughly 300 seconds and 360 meters long. Fig.~\ref{fig:maps} shows the path overlaid on a map as well as example views along the path.

{\em  Data acquisition:} We captured the sensor data from the iPad using an app developed for \cite{cortes2018advio}. It records the acceleration, rotation rate, magnetometer readings, barometric pressure, platform locations, video frames, and ARKit poses (see below). The multiple streams are time-synchronized on the device. The GNSS based locations are recorded trough Apple's platform location API (CoreLocation, in high accuracy mode) which merges satellite and ground based global positioning systems. Each location has an uncertainty radius associated and it reflects the source and quality of the position estimation. In general, satellite based locations are the most accurate, Wi-Fi based locations are less accurate, and cell tower based locations are the least accurate. The platform location service has several modes of operation that let the developer pick a balance between accuracy (and frequency) and energy usage. The WGS coordinates provided by the platform/GNSS were transformed to a metric East-North-Up (ENU) coordinate system before fusing with the model. The accelerometer, gyroscope, magnetometer, and barometer were captured and stored in their raw form. Processing of the data was in this case done off-line.

{\em Visual-inertial reference track:} For reference, we captured the built-in visual-inertial  odometry (Apple ARKit API) poses. ARKit is designed for augmented reality and thus, it is very accurate in the short-term. However, it has a tendency to drift if the same scene is not seen again (no visual loop-closures). Each pose (acquired at 60~Hz) consists of a translation in meters and a unit quaternion which represents the rotation. These values are used to construct the local-to-global transformation of the coordinate frame in each sample.

In these case, the ARKit systems keeps a very accurate track of the straight walks. This is due to the features remaining in the frame for long. The corners are where the system loses most of its tracked features. Then, small errors on the angle of a corner accumulate and by the end of the capture there is a few meters of drift. This also be seen in Fig.~\ref{fig:map-arkit}.

{\em Optimal track:} If the system described in Sec.~\ref{sec:methods} is used with all the available information, the result will not have room to improve with iteration. This is due to the dense position updates that correct any drift caused by linearization errors. This track is a good baseline for the `ideal' track. It has short-term accuracy comparable with ARKit and long-term drift correction due to the GNSS signal.

\begin{figure*}[t!]
  \begin{subfigure}{.48\textwidth}
  \centering\scriptsize
  \setlength{\figurewidth}{0.9\columnwidth}
  \setlength{\figureheight}{0.55\figurewidth}
%
%
%
\begin{tikzpicture}

\begin{axis}[%
width=0.951\figurewidth,
height=\figureheight,
at={(0\figurewidth,0\figureheight)},
scale only axis,
unbounded coords=jump,
xmin=0,
grid=major,
xmax=150,
xlabel style={font=\color{white!15!black}},
xlabel={Scale (seconds)},
ymin=0,
ymax=3.5,
ylabel style={font=\color{white!15!black}},
ylabel={RMSE (meters)},
axis background/.style={fill=white},
title style={font=\bfseries},
title={SARMSE},
legend style={at={(0.03,0.97)}, anchor=north west, legend cell align=left, align=left, draw=white!15!black}
]

\addplot [color=mycolor1, line width=2.0pt]
  table[row sep=crcr]{%
1	0.0295666752865756\\
1.11	0.0328792653451598\\
1.23	0.0365298663198971\\
1.37	0.0408112655035204\\
1.51	0.0451087705145638\\
1.68	0.0503053563938526\\
1.86	0.0558006834742257\\
2.07	0.0621727759010861\\
2.29	0.0689897350910519\\
2.54	0.076547159170273\\
2.82	0.085067657609482\\
3.13	0.0941469279843259\\
3.47	0.104356740515995\\
3.85	0.115774639198117\\
4.27	0.12869113632169\\
4.74	0.143267024808647\\
5.26	0.159497174951616\\
5.83	0.17716122608223\\
6.47	0.197383534378912\\
7.18	0.219976333691247\\
7.96	0.245034054886429\\
8.83	0.273410498277426\\
9.8	0.305975825061549\\
10.87	0.342333005595555\\
12.06	0.38388326541002\\
13.38	0.431363448564266\\
14.84	0.485295348953181\\
16.46	0.547047579404793\\
18.26	0.617627174930657\\
20.26	0.698141592918734\\
22.47	0.789144336408466\\
24.93	0.89260645329516\\
27.66	1.00996733385768\\
30.68	1.14248516587362\\
34.03	1.29166486574071\\
37.75	1.45645424455509\\
41.88	1.63422407651376\\
46.46	1.82059433244387\\
51.54	2.00991836544386\\
57.17	2.20299099117563\\
63.42	2.40575604878816\\
70.35	2.61887602203737\\
78.05	2.83487968373879\\
86.58	3.03934480822932\\
96.04	3.21061884392895\\
106.54	3.31137514269755\\
118.19	3.31826913482425\\
131.11	3.27075252280595\\
145.44	3.17053767855753\\
161.34	nan\\
};
\addlegendentry{EKF}

\addplot [color=white!10!black, line width=2.0pt]
  table[row sep=crcr]{%
1	0.031394475707068\\
1.11	0.0349070552725196\\
1.23	0.0387777246306298\\
1.37	0.0433005907214807\\
1.51	0.0478405973475794\\
1.68	0.0533939733109974\\
1.86	0.059314212921918\\
2.07	0.066270955809566\\
2.29	0.0735658412349426\\
2.54	0.0819338602682801\\
2.82	0.0913686227292227\\
3.13	0.101901976373759\\
3.47	0.113624564119966\\
3.85	0.126917446453959\\
4.27	0.141824701516599\\
4.74	0.158550807452906\\
5.26	0.177161807373028\\
5.83	0.197647432555559\\
6.47	0.220966890223834\\
7.18	0.247039910050171\\
7.96	0.275901361050138\\
8.83	0.308388108233236\\
9.8	0.345201360728372\\
10.87	0.386026244611005\\
12.06	0.431961186536049\\
13.38	0.483396176319689\\
14.84	0.540532308127016\\
16.46	0.604193952248095\\
18.26	0.674847716699848\\
20.26	0.752884270655522\\
22.47	0.837983861420693\\
24.93	0.930913298615991\\
27.66	1.0313768887644\\
30.68	1.13965671808991\\
34.03	1.25693567634348\\
37.75	1.38208533663572\\
41.88	1.51384579592073\\
46.46	1.64788573376998\\
51.54	1.77274498963092\\
57.17	1.88770226106089\\
63.42	2.00569170385537\\
70.35	2.14374792043648\\
78.05	2.28664866149513\\
86.58	2.37756219623573\\
96.04	2.44244397818007\\
106.54	2.45442810278548\\
118.19	2.38761486126797\\
131.11	2.30706425872593\\
145.44	2.23164950355863\\
161.34	nan\\
};
\addlegendentry{GIEKF-20}

\addplot [color=mycolor2, line width=2.0pt]
  table[row sep=crcr]{%
1	0.130454133477625\\
1.11	0.14465473615337\\
1.23	0.16021731623053\\
1.37	0.178036122993361\\
1.51	0.195570817495884\\
1.68	0.216488443043983\\
1.86	0.237987548666439\\
2.07	0.262975200285092\\
2.29	0.288649451003101\\
2.54	0.316959409404989\\
2.82	0.347683226205139\\
3.13	0.380384405000139\\
3.47	0.415270592768225\\
3.85	0.452766713555367\\
4.27	0.492953618998951\\
4.74	0.536254576316314\\
5.26	0.582298615229828\\
5.83	0.630418751373987\\
6.47	0.682087289862019\\
7.18	0.736758276145713\\
7.96	0.794594984577238\\
8.83	0.857144396350219\\
9.8	0.92500872658287\\
10.87	0.996514490585317\\
12.06	1.07178760986464\\
13.38	1.14947276848169\\
14.84	1.22848433552429\\
16.46	1.30775386567047\\
18.26	1.38691716084105\\
20.26	1.4629035334623\\
22.47	1.53266349909241\\
24.93	1.59939072472064\\
27.66	1.6623668884737\\
30.68	1.72117529134287\\
34.03	1.77846895495077\\
37.75	1.83515280358987\\
41.88	1.88782032055415\\
46.46	1.93068881194601\\
51.54	1.96903347054924\\
57.17	2.00455235011528\\
63.42	2.03603316473206\\
70.35	2.05445816796947\\
78.05	2.06951677869069\\
86.58	2.07478969570054\\
96.04	2.09061711743485\\
106.54	2.11865909819888\\
118.19	2.11326239948384\\
131.11	2.07156033707535\\
145.44	2.04413207353167\\
161.34	nan\\
};
\addlegendentry{GNSS}

\addplot [color=mycolor3, line width=2.0pt]
  table[row sep=crcr]{%
1	0.0352079367048368\\
1.11	0.0389594024374723\\
1.23	0.0430132033848941\\
1.37	0.047827592411805\\
1.51	0.0526543440940584\\
1.68	0.0585022077374384\\
1.86	0.0647136116951427\\
2.07	0.0718481533571041\\
2.29	0.0793965542331155\\
2.54	0.0879436621016487\\
2.82	0.0974832052889808\\
3.13	0.10791471140301\\
3.47	0.11947056180855\\
3.85	0.132483758757824\\
4.27	0.146969979730993\\
4.74	0.162882510982823\\
5.26	0.180451965136541\\
5.83	0.199625522398595\\
6.47	0.221378187797825\\
7.18	0.2455097075229\\
7.96	0.272047791731009\\
8.83	0.301757735054073\\
9.8	0.335276342282115\\
10.87	0.372213128912039\\
12.06	0.413548409601032\\
13.38	0.459600058276931\\
14.84	0.510436078619948\\
16.46	0.566663540686881\\
18.26	0.628679765418338\\
20.26	0.696965753445735\\
22.47	0.771564786372881\\
24.93	0.853393237329687\\
27.66	0.942979019361998\\
30.68	1.04011210362684\\
34.03	1.14306095790385\\
37.75	1.24904840731191\\
41.88	1.35509194500463\\
46.46	1.45036850557227\\
51.54	1.52820084747275\\
57.17	1.58733652944553\\
63.42	1.63281171339051\\
70.35	1.66663935854061\\
78.05	1.69559416148648\\
86.58	1.72259413654807\\
96.04	1.76284638346968\\
106.54	1.85807840804763\\
118.19	2.05290534143434\\
131.11	2.28524769889152\\
145.44	2.4728239533063\\
161.34	nan\\
};
\addlegendentry{ARKit}

\end{axis}
\end{tikzpicture}%
  \caption{SARMSE for gap track} 
  \label{fig:gap_error}
  \end{subfigure}
  \hfill
  \begin{subfigure}{.48\textwidth}  
  \centering\scriptsize
  \setlength{\figurewidth}{0.9\columnwidth}
  \setlength{\figureheight}{0.55\figurewidth}
%
%
%
\begin{tikzpicture}

\begin{axis}[%
width=0.951\figurewidth,
height=\figureheight,
at={(0\figurewidth,0\figureheight)},
scale only axis,
unbounded coords=jump,
xmin=0,
xmax=150,
grid=major,
xlabel style={font=\color{white!15!black}},
xlabel={Scale (seconds)},
ymin=0,
ymax=3.5,
ylabel style={font=\color{white!15!black}},
ylabel={RMSE (meters)},
axis background/.style={fill=white},
title style={font=\bfseries},
title={SARMSE},
legend style={at={(0.03,0.97)}, anchor=north west, legend cell align=left, align=left, draw=white!15!black}
]
\addplot [color=mycolor1, line width=2.0pt]
  table[row sep=crcr]{%
1	0.032017194382645\\
1.11	0.03561211332296\\
1.23	0.0395486922113737\\
1.37	0.044171939829432\\
1.51	0.0488220382916695\\
1.68	0.0544887549142748\\
1.86	0.0604935711346532\\
2.07	0.0674346602926045\\
2.29	0.0747981072009472\\
2.54	0.083202323372887\\
2.82	0.0927057459313173\\
3.13	0.103197412688599\\
3.47	0.11487884183033\\
3.85	0.128077265159054\\
4.27	0.14297499373578\\
4.74	0.159860841819991\\
5.26	0.17868641666642\\
5.83	0.199370337591504\\
6.47	0.222859277673014\\
7.18	0.249074408566408\\
7.96	0.278026181977096\\
8.83	0.310502935387584\\
9.8	0.347133474363049\\
10.87	0.387419788772586\\
12.06	0.432331506855057\\
13.38	0.482085697621047\\
14.84	0.53654079634592\\
16.46	0.596240529648615\\
18.26	0.661192207822453\\
20.26	0.73145086032192\\
22.47	0.806702084390093\\
24.93	0.88791441456556\\
27.66	0.975187795376343\\
30.68	1.06791759543301\\
34.03	1.16516611333544\\
37.75	1.26422582434774\\
41.88	1.36316245655558\\
46.46	1.45847447558942\\
51.54	1.53459993251866\\
57.17	1.58157103184548\\
63.42	1.59993604792294\\
70.35	1.59962994421406\\
78.05	1.60833917356807\\
86.58	1.63599042712205\\
96.04	1.68404081977879\\
106.54	1.71739261967399\\
118.19	1.71894323744209\\
131.11	1.70963049342741\\
145.44	1.68731965026399\\
161.34	nan\\
};
\addlegendentry{EKF}

\addplot [color=white!10!black, line width=2.0pt]
  table[row sep=crcr]{%
1	0.0208185011057512\\
1.11	0.0231245038826318\\
1.23	0.0256241252519825\\
1.37	0.0286102234433439\\
1.51	0.0315926449805627\\
1.68	0.0352186587558815\\
1.86	0.0390674359049684\\
2.07	0.043489261449013\\
2.29	0.0481341624971531\\
2.54	0.0533776102158328\\
2.82	0.059258883421733\\
3.13	0.0656261948641016\\
3.47	0.0727410402288072\\
3.85	0.0807621300478441\\
4.27	0.0897817776243718\\
4.74	0.0997053997404724\\
5.26	0.110674158289654\\
5.83	0.122628597518188\\
6.47	0.136283592795395\\
7.18	0.151459288616762\\
7.96	0.168176455959192\\
8.83	0.186941787109788\\
9.8	0.208259405251184\\
10.87	0.231706581330212\\
12.06	0.258044392315053\\
13.38	0.287540938301343\\
14.84	0.320255889108423\\
16.46	0.356852431448761\\
18.26	0.397560784107311\\
20.26	0.442668814748297\\
22.47	0.492138320922857\\
24.93	0.546469898826856\\
27.66	0.605325504885005\\
30.68	0.668102687855488\\
34.03	0.73485658627738\\
37.75	0.805187476743621\\
41.88	0.877390243911658\\
46.46	0.948076652314052\\
51.54	1.01351332850109\\
57.17	1.07389720060193\\
63.42	1.12837478337444\\
70.35	1.1685380083063\\
78.05	1.19450901730807\\
86.58	1.2228059088861\\
96.04	1.26752999586797\\
106.54	1.33172974982478\\
118.19	1.34432073953336\\
131.11	1.30439379336572\\
145.44	1.26824864370474\\
161.34	nan\\
};
\addlegendentry{GIEKF-20}

\addplot [color=mycolor2, line width=2.0pt]
  table[row sep=crcr]{%
1	0.130454133477625\\
1.11	0.14465473615337\\
1.23	0.16021731623053\\
1.37	0.178036122993361\\
1.51	0.195570817495884\\
1.68	0.216488443043983\\
1.86	0.237987548666439\\
2.07	0.262975200285092\\
2.29	0.288649451003101\\
2.54	0.316959409404989\\
2.82	0.347683226205139\\
3.13	0.380384405000139\\
3.47	0.415270592768225\\
3.85	0.452766713555367\\
4.27	0.492953618998951\\
4.74	0.536254576316314\\
5.26	0.582298615229828\\
5.83	0.630418751373987\\
6.47	0.682087289862019\\
7.18	0.736758276145713\\
7.96	0.794594984577238\\
8.83	0.857144396350219\\
9.8	0.92500872658287\\
10.87	0.996514490585317\\
12.06	1.07178760986464\\
13.38	1.14947276848169\\
14.84	1.22848433552429\\
16.46	1.30775386567047\\
18.26	1.38691716084105\\
20.26	1.4629035334623\\
22.47	1.53266349909241\\
24.93	1.59939072472064\\
27.66	1.6623668884737\\
30.68	1.72117529134287\\
34.03	1.77846895495077\\
37.75	1.83515280358987\\
41.88	1.88782032055415\\
46.46	1.93068881194601\\
51.54	1.96903347054924\\
57.17	2.00455235011528\\
63.42	2.03603316473206\\
70.35	2.05445816796947\\
78.05	2.06951677869069\\
86.58	2.07478969570054\\
96.04	2.09061711743485\\
106.54	2.11865909819888\\
118.19	2.11326239948384\\
131.11	2.07156033707535\\
145.44	2.04413207353167\\
161.34	nan\\
};
\addlegendentry{GNSS}

\addplot [color=mycolor3, line width=2.0pt]
  table[row sep=crcr]{%
1	0.0352079367048368\\
1.11	0.0389594024374723\\
1.23	0.0430132033848941\\
1.37	0.047827592411805\\
1.51	0.0526543440940584\\
1.68	0.0585022077374384\\
1.86	0.0647136116951427\\
2.07	0.0718481533571041\\
2.29	0.0793965542331155\\
2.54	0.0879436621016487\\
2.82	0.0974832052889808\\
3.13	0.10791471140301\\
3.47	0.11947056180855\\
3.85	0.132483758757824\\
4.27	0.146969979730993\\
4.74	0.162882510982823\\
5.26	0.180451965136541\\
5.83	0.199625522398595\\
6.47	0.221378187797825\\
7.18	0.2455097075229\\
7.96	0.272047791731009\\
8.83	0.301757735054073\\
9.8	0.335276342282115\\
10.87	0.372213128912039\\
12.06	0.413548409601032\\
13.38	0.459600058276931\\
14.84	0.510436078619948\\
16.46	0.566663540686881\\
18.26	0.628679765418338\\
20.26	0.696965753445735\\
22.47	0.771564786372881\\
24.93	0.853393237329687\\
27.66	0.942979019361998\\
30.68	1.04011210362684\\
34.03	1.14306095790385\\
37.75	1.24904840731191\\
41.88	1.35509194500463\\
46.46	1.45036850557227\\
51.54	1.52820084747275\\
57.17	1.58733652944553\\
63.42	1.63281171339051\\
70.35	1.66663935854061\\
78.05	1.69559416148648\\
86.58	1.72259413654807\\
96.04	1.76284638346968\\
106.54	1.85807840804763\\
118.19	2.05290534143434\\
131.11	2.28524769889152\\
145.44	2.4728239533063\\
161.34	nan\\
};
\addlegendentry{ARKit}

\end{axis}
\end{tikzpicture}%
  \caption{SARMSE for subsampled track}
  \label{fig:skip_error}  
  \end{subfigure}  
  \caption{The iteration improves the estimation, specially in the large scales. In (a), the gap is too big (160~m) to allow improvement at the lower scales. In (b), the evenly but sparsely sampled GNSS help the iterated scheme improve more. The error is with respect to the full track reconstruction that uses all data.}
\end{figure*}
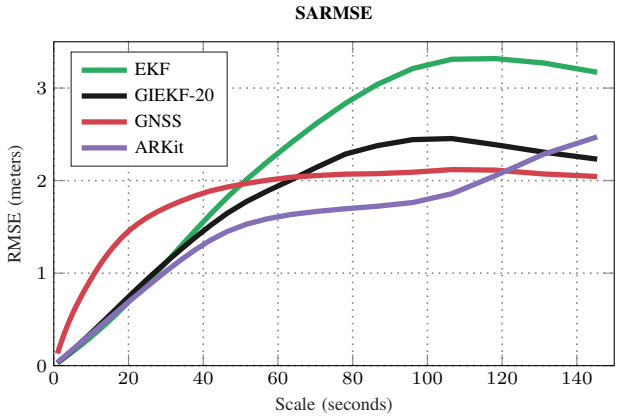
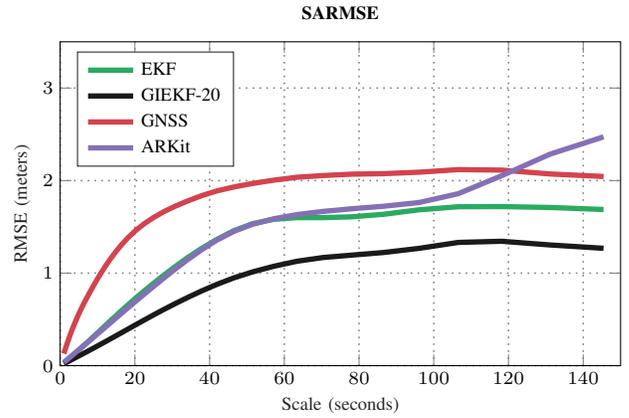

\begin{figure}[!t]
  \setlength{\figurewidth}{.9\columnwidth}
  \setlength{\figureheight}{0.24\figurewidth}
  \pgfplotsset{yticklabel style={rotate=90}, ylabel style={yshift=0pt},clip=true,scale only axis,axis on top,clip marker paths,legend style={row sep=0pt},xlabel near ticks,legend style={fill=white}}
  \begin{subfigure}[b]{\columnwidth}
    \scriptsize

%
%
\definecolor{mycolor1}{rgb}{0.26667,0.44706,0.70980}%
\begin{tikzpicture}

\begin{axis}[%
xmin=0,
xmax=20,
xtick={0,2,4,6,8,10,12,14,16,18,20},
xticklabels={\empty},
xlabel={\phantom{Iteration}},
xmajorgrids,
ymin=15,
ymax=21,
ylabel={RMSE},
ymajorgrids,
axis background/.style={fill=white},
legend style={legend cell align=left,align=left,draw=white!15!black},
height=\figureheight,
width=\figurewidth,
scale only axis
]
\addplot [color=mycolor1,solid,line width=1.5pt,forget plot]
  table[row sep=crcr]{%
1	20.8946136682223\\
2	20.1793359891757\\
3	19.382239152607\\
4	18.6475162910548\\
5	17.9730745925118\\
6	17.3492746280452\\
7	16.8794033498084\\
8	16.5459134930383\\
9	16.3317576992043\\
10	16.2108578373622\\
11	16.1412693865164\\
12	16.0985899498145\\
13	16.0723393825709\\
14	16.0554539058291\\
15	16.0407799607658\\
16	16.0231990560509\\
17	15.9983619264038\\
18	15.9632857677363\\
19	15.9212366229115\\
20	15.8752460463637\\
};
\end{axis}
\end{tikzpicture}%
  \end{subfigure}    
  \begin{subfigure}[b]{\columnwidth}
    \scriptsize

%
%
\definecolor{mycolor1}{rgb}{0.16471,0.67059,0.38039}%
\begin{tikzpicture}

\begin{axis}[%
xmin=0,
xmax=20,
xtick={0,2,4,6,8,10,12,14,16,18,20},
xticklabels={\empty},
xlabel={\phantom{Iteration}},
xmajorgrids,
ymin=13,
ymax=20,
ytick={14, 16, 18},
ylabel={MAE},
ymajorgrids,
axis background/.style={fill=white},
legend style={legend cell align=left,align=left,draw=white!15!black},
height=\figureheight,
width=\figurewidth,
scale only axis
]
\addplot [color=mycolor1,solid,line width=1.5pt,forget plot]
  table[row sep=crcr]{%
1	19.7807685483227\\
2	18.9661502995215\\
3	18.0443100387973\\
4	17.1565758282642\\
5	16.3022757177017\\
6	15.5547669041366\\
7	15.0004882859512\\
8	14.6328916279788\\
9	14.4267693938834\\
10	14.2902219991037\\
11	14.1956492294932\\
12	14.1317036691165\\
13	14.0788291510151\\
14	14.0367293364351\\
15	14.0265988128793\\
16	14.0368945019322\\
17	14.0409218174346\\
18	14.0313964690462\\
19	14.0130653494244\\
20	13.9869789177196\\
};
\end{axis}
\end{tikzpicture}%
  \end{subfigure}    
  \begin{subfigure}[b]{\columnwidth}
    \scriptsize

%
%
\definecolor{mycolor1}{rgb}{0.82745,0.26275,0.30588}%
\begin{tikzpicture}

\begin{axis}[%
xmin=0,
xmax=20,
xlabel={Iteration},
xmajorgrids,
ymin=500,
ymax=850,
ylabel={NLPD},
ymajorgrids,
axis background/.style={fill=white},
legend style={legend cell align=left,align=left,draw=white!15!black},
height=\figureheight,
width=\figurewidth,
scale only axis
]
\addplot [color=mycolor1,solid,line width=1.5pt,forget plot]
  table[row sep=crcr]{%
1	806.973527305857\\
2	756.965007682409\\
3	704.436725741049\\
4	663.469043233259\\
5	630.644188801404\\
6	602.121663452683\\
7	581.398022212637\\
8	566.711762378849\\
9	556.301393382596\\
10	548.796126364928\\
11	543.06683508414\\
12	538.134003914932\\
13	533.759647054474\\
14	529.640070068938\\
15	525.48817432835\\
16	521.113373683173\\
17	516.3881702825\\
18	511.271135022325\\
19	505.939545023301\\
20	500.506220905549\\
};
\end{axis}
\end{tikzpicture}%
  \end{subfigure}    

  \caption{Progression of various error metrics over the course of global iterated EKF iteration steps. Top subfigure depicts root mean square error (\textcolor{mycolor0}{RMSE}), the middle subfigure mean absolute error (\textcolor{mycolor1}{MAE}), and the bottom negative log predictive density (\textcolor{mycolor2}{NLPD})---all w.r.t.\ left-out GNSS locations. After 20 iterations the RMSE/MAE which capture the mean seem converged, but NLPD still shows a slope.}
  \label{fig:skip}

\end{figure}
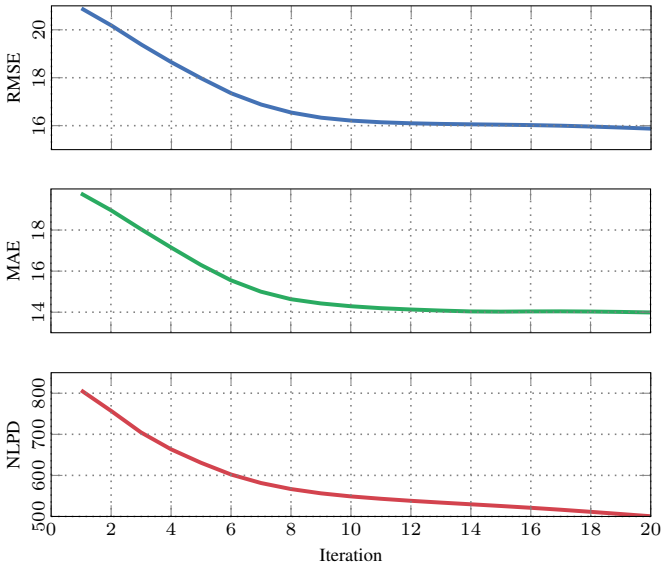

\subsection{GNSS gap and subsampling experiments}
\noindent
In order to demonstrate the drift correction caused by the iteration a gap is artificially created in the GNSS signal. In practice this is a common occurrence that happens when the user goes inside a structure or underground. The gap is 40\% of the capture time, roughly 113 seconds or 160 meters of walking. The GNSS signal gap is shown in Fig.~\ref{fig:map-gnss} (the green GNSS markers).

The iterative scheme is most useful when the state variables do not change. In this case the bias of the sensors does not change. This directly influences the linearization point for the orientation update and keeps a better estimate of the orientation.  Fig.~\ref{fig:map-ours} shows how this translates into a more `square' track that matches the street map over the iterations. The first iteration (`EKF') corresponds to the non-iterated method by \cite{Solin+Cortes+Rahtu+Kannala:2018-FUSION}. Fig.~\ref{fig:gap_error} shows the SARMSE results for this track and experiment. Clearly showing that after 20 global iterations, the global scale has clearly improved, which is also visible in Fig.~\ref{fig:map-ours}.

Another ablation of the data capture consists of sampling GNSS locations sparsely (only using $\nicefrac{1}{35}$ of the GNSS locations). The GIEKF scheme can be seen as very beneficial in this case, improving the SARMSE clearly over the course of iterations (Fig.~\ref{fig:skip_error}). Fig.~\ref{fig:skip} shows the evolution of various error metrics over the course of iterations---with respect to left-out GNSS locations.

\subsection{Quantitative runs on the ADVIO data set}
\noindent
ADVIO~\cite{cortes2018advio} is a public data set for long-range visual inertial odometry for hand-held devices, which contains sequences in both indoor and outdoor environments. The ground-truth is captured by combining IMU data with the set of fixation points that consists of time--position pairs. The location fixes are manually determined according to a reference video that views the recorder and precise building floor plans. This dataset has plenty of environments and different motion modes.

The ground-truth tracks are sampled every three seconds, and the sampled planar position is used as a GNSS update (with added jitter). The height of each sampled point is used as the barometer update with a standard deviation of $0.2$~meters. The sub-sampled track is linearly interpolated and the distance to the ground-truth is also measured. The resulting RMSE with respect to the ground-truth is shown in Table~\ref{tbl:results}.

\section{Discussion and Conclusion}
\noindent
Accurate estimation of the sensor biases is crucial for correct odometry. Online estimation in an augmented state has proven to perform very well when enough position and velocity information is available (fix points or ZUPTs). However, gaps in this information make the linearization errors of the EKF much more pronounced. Iterative schemes allow the linearization point to be improved and thus boost the performance of the odometry system.

The model is somewhat sensitive to the parameter choices and initial state covariances for the model biases.

Alternative strategies for non-linear filtering, such as sigma-point methods (unscented Kalman filters, see \cite{Sarkka:2013}) or sequential Monte Carlo methdos (particle filters), were not considered in this paper. Apart from being fast, the EKF is also known to perform well for models like this where local linearization is a desirable approximation. However, performing some of the updates with sigma-point methods, and running a particle filter for validation, would be an interesting avenue to explore.

In our experiments, the SARMSE metric shows how the iteration improves the accuracy of the estimated track in the medium and long-range (Fig.~\ref{fig:gap_error}). The global iteration works best in the parts of the system that do not change, in this case, the sensor biases in the state. The ADVIO experiment shows how the system improves the tracks when random gaps are created. Even with the high quality (manually annotated) position updates, the estimation improves with iteration. 

The improvement depends on a good enough initial estimate. The iteration improves tracks that already have a reasonable approximation of the motion. Once the estimation has diverged, the iteration usually makes it worse. The model is designed to deal with gaps in GNSS data. Outlier detection and managing  faulty GNSS signals is not adressed. However, There has been a lot of work into this subject and clould be integrated with the proposed system.

\begin{table}[!tb]
  \caption{Median error (meters) of the ADVIO sequences.}
  \label{tbl:results}
  \centering
  \begin{filecontents*}{fig/advio.csv}
1,0.329,0.273,2.171,9.847
2,0.257,0.181,2.241,1.374
3,0.926,0.389,1.150,1.324
4,0.921,0.372,3.722,10.32
5,0.601,0.290,1.476,4.896
6,0.275,0.254,3.748,1.504
7,0.574,0.526,3.004,10.0
8,0.387,0.358,1.110,2.081
9,0.731,0.371,2.564,2.056
10,0.222,0.186,1.579,1.335
11,0.226,0.215,2.548,1.298
12,0.493,0.215,1.505,1.414
13,0.289,0.264,1.162,2.130
14,0.325,0.310,1.253,3.090
15,0.779,0.324,0.802,1.075
16,0.829,0.239,1.135,4.073
17,0.257,0.252,1.528,2.063
18,0.406,0.254,0.561,2.683
19,0.353,0.410,0.842,2.245
20,0.300,0.219,19.162,1.631
21,0.228,0.155,16.678,1.590
22,0.211,0.221,4.151,1.500
23,0.533,0.724,9440,1.493
\end{filecontents*}
\pgfplotstabletypeset[
    col sep=comma,
    string type,
    every head row/.style={%
    output empty row,
        before row={\hline
             Sequence & EKF & GIEKF-20 & ARKit  & Line interpolation  \\
        },
        after row=\hline
    },
    every last row/.style={ 
    after row={\hline
             Mean & 0.454 & 0.304 & 413.6 & 3.081\\       
             Median & 0.353 & 0.264 & 1.579 &2.055\\    
        }
    },
    ]{fig/advio.csv}
\end{table}

Further material related to this paper is available online:
\url{https://aaltoml.github.io/iterated-INS}

\section*{Acknowledgments}
\noindent
This research was supported by the Academy of Finland grants 308640, 277685, and 295081. We  acknowledge the computational resources provided by the Aalto Science-IT project.

{\small
\bibliographystyle{ieee}
\bibliography{bibliography}

\begin{thebibliography}{10}\itemsep=-1pt

\bibitem{OXTS}
{Oxford Technical Solutions~Ltd.~[Accessed March 15, 2019]}.
\newblock \url{https://www.oxts.com/products/rt3000/}.

\bibitem{Pix4d}
{Pix4D~S.A.~[Accessed March 15, 2019]}.
\newblock \url{https://www.pix4d.com/}.

\bibitem{Bar-Shalom+Li+Kirubarajan:2001}
Y.~Bar-Shalom, X.-R. Li, and T.~Kirubarajan.
\newblock {\em Estimation with Applications to Tracking and Navigation}.
\newblock Wiley-Interscience, New York, 2001.

\bibitem{Bloesch+Omari+Hutter+Siegwart:2015}
M.~Bloesch, S.~Omari, M.~Hutter, and R.~Y. Siegwart.
\newblock Robust visual inertial odometry using a direct {EKF}-based approach.
\newblock In {\em Proceedings of IROS}, pages 298--304, 2015.

\bibitem{Britting:2010}
K.~R. Britting.
\newblock {\em Inertial Navigation Systems Analysis}.
\newblock Wiley-Interscience, New York, 2010.

\bibitem{Trigoni}
C.~Chen, X.~Lu, A.~Markham, and N.~Trigoni.
\newblock {IONet}: Learning to cure the curse of drift in inertial odometry.
\newblock pages 6468--6476, 2018.

\bibitem{Chen+Meng+Wang+Zhang+Tian+Yang:2015}
G.~Chen, X.~Meng, Y.~Wang, Y.~Zhang, P.~Tian, and H.~Yang.
\newblock Integrated {WiFi}/{PDR}/smartphone using an unscented {K}alman filter
  algorithm for {3D} indoor localization.
\newblock {\em Sensors}, 15(9):24595--24614, 2015.

\bibitem{cortes2018mlsp}
S.~Cort{\'e}s, A.~Solin, and J.~Kannala.
\newblock Deep learning based speed estimation for constraining strapdown
  inertial navigation on smartphones.
\newblock In {\em Proceedings of MLSP}, 2018.

\bibitem{cortes2018advio}
S.~Cort{\'e}s, A.~Solin, E.~Rahtu, and J.~Kannala.
\newblock {ADVIO}: {A}n authentic dataset for visual-inertial odometry.
\newblock In {\em Proceedings of ECCV}, pages 419--434, 2018.

\bibitem{Foxlin:2005}
E.~Foxlin.
\newblock Pedestrian tracking with shoe-mounted inertial sensors.
\newblock {\em Computer Graphics and Applications}, 25(6):38--46, 2005.

\bibitem{Gelman_et_al:2013}
A.~Gelman, J.~B. Carlin, H.~S. Stern, D.~B. Dunson, A.~Vehtari, and D.~B.
  Rubin.
\newblock {\em Bayesian Data Analysis}.
\newblock Chapman and Hall/CRC, Boca Raton, FL, third edition, 2013.

\bibitem{Harle:2013}
R.~Harle.
\newblock A survey of indoor inertial positioning systems for pedestrians.
\newblock {\em Communications Surveys \& Tutorials}, 15(3):1281--1293, 2013.

\bibitem{Hesch+Kottas+Bowman+Roumeliotis:2014}
J.~A. Hesch, D.~G. Kottas, S.~L. Bowman, and S.~I. Roumeliotis.
\newblock Consistency analysis and improvement of vision-aided inertial
  navigation.
\newblock {\em Transactions on Robotics}, 30(1):158--176, 2014.

\bibitem{Jekeli:2001}
C.~Jekeli.
\newblock {\em Inertial Navigation Systems with Geodetic Applications}.
\newblock Walter de Gruyter, Berlin, Germany, 2001.

\bibitem{Kang+Han:2015}
W.~Kang and Y.~Han.
\newblock {SmartPDR}: {S}martphone-based pedestrian dead reckoning for indoor
  localization.
\newblock {\em Sensors}, 15(5):2906--2916, 2015.

\bibitem{Kok:2015-newton}
M.~Kok, J.~Dahlin, T.~Sch{\"o}n, and A.~Wills.
\newblock Newton-based maximum likelihood estimation in nonlinear state space
  models.
\newblock In {\em Proceedings of the 17th IFAC Symposium on System
  Identification (SYSID)}, pages 398--403, 2015.

\bibitem{Li+Kim+Mourikis:2013}
M.~Li, B.~Kim, and A.~Mourikis.
\newblock Real-time motion tracking on a cellphone using inertial sensing and a
  rolling shutter camera.
\newblock In {\em Proceedings of ICRA}, pages 4712--4719, 2013.

\bibitem{Maybeck:1982}
P.~S. Maybeck.
\newblock {\em Stochastic Models, Estimation and Control}, volume~2.
\newblock Academic Press, New York, NY, 1982.

\bibitem{Menke+Zakhor:2015}
J.~Menke and A.~Zakhor.
\newblock Multi-modal indoor positioning of mobile devices.
\newblock In {\em Proceedings of the International Conference on Indoor
  Positioning and Indoor Navigation (IPIN)}, 2015.

\bibitem{Nilsson+Gupta+Handel:2014}
J.-O. Nilsson, A.~K. Gupta, and P.~H\"andel.
\newblock Foot-mounted inertial navigation made easy.
\newblock In {\em Proceedings of IPIN}, pages 24--29, 2014.

\bibitem{Nilsson+Zachariah+Skog+Handel:2013}
J.-O. Nilsson, D.~Zachariah, I.~Skog, and P.~H{\"a}ndel.
\newblock Cooperative localization by dual foot-mounted inertial sensors and
  inter-agent ranging.
\newblock {\em Journal on Advances in Signal Processing}, 2013(1):1--17, 2013.

\bibitem{Renaudin+Combettes:2014}
V.~Renaudin and C.~Combettes.
\newblock Magnetic, acceleration fields and gyroscope quaternion
  ({MAGYQ})-based attitude estimation with smartphone sensors for indoor
  pedestrian navigation.
\newblock {\em Sensors}, 14(12):22864--22890, 2014.

\bibitem{Sarkka:2013}
S.~S{\"a}rkk{\"a}.
\newblock {\em Bayesian Filtering and Smoothing}.
\newblock Cambridge University Press, 2013.

\bibitem{Solin+Cortes+Rahtu+Kannala:2018-FUSION}
A.~Solin, S.~Cortes, E.~Rahtu, and J.~Kannala.
\newblock Inertial odometry on handheld smartphones.
\newblock In {\em Proceedings of FUSION}, 2018.

\bibitem{Solin+Cortes+Rahtu+Kannala:2018-WACV}
A.~Solin, S.~Cortes, E.~Rahtu, and J.~Kannala.
\newblock {PIVO}: {P}robabilistic inertial-visual odometry for occlusion-robust
  navigation.
\newblock In {\em Proceedings of WACV}, pages 616--625, 2018.

\bibitem{Solin+Sarkka+Kannala+Rahtu:2016}
A.~Solin, S.~S{\"a}rkk{\"a}, J.~Kannala, and E.~Rahtu.
\newblock Terrain navigation in the magnetic landscape: {P}article filtering
  for indoor positioning.
\newblock In {\em Proceedings of ENC}, pages 1--9, 2016.

\bibitem{thrun_et_al:2005}
S.~Thrun, W.~Burgard, and D.~Fox.
\newblock {\em Probabilistic robotics}.
\newblock MIT press, 2005.

\bibitem{Titterton+Weston:2004}
D.~H. Titterton and J.~L. Weston.
\newblock {\em Strapdown Inertial Navigation Technology}.
\newblock The Institution of Electrical Engineers, 2004.

\bibitem{Woodman:2010}
O.~J. Woodman.
\newblock {\em Pedestrian Localisation for Indoor Environments}.
\newblock PhD thesis, University of Cambridge, Cambridge, UK, September 2010.

\bibitem{Xiao+Wen+Markham+Trigoni:2014}
Z.~Xiao, H.~Wen, A.~Markham, and N.~Trigoni.
\newblock Robust pedestrian dead reckoning ({R-PDR}) for arbitrary mobile
  device placement.
\newblock In {\em Proceedings of IPIN}, pages 187--196, 2014.

\bibitem{Yan+Shan+Furukawa:2018}
H.~Yan, Q.~Shan, and Y.~Furukawa.
\newblock {RIDI}: {R}obust {IMU} double integration.
\newblock In {\em Proceedings of ECCV}, 2018.

\bibitem{Yuan+Yu+Zhang+Wang+Liu:2015}
X.~Yuan, S.~Yu, S.~Zhang, G.~Wang, and S.~Liu.
\newblock Quaternion-based unscented {K}alman filter for accurate indoor
  heading estimation using wearable multi-sensor system.
\newblock {\em Sensors}, 15(5):10872--10890, 2015.

\bibitem{zhang+Scaramuzza:2018}
Z.~Zhang and D.~Scaramuzza.
\newblock A tutorial on quantitative trajectory evaluation for visual
  (-inertial) odometry.
\newblock In {\em Proceedings of IROS}, pages 7244--7251. IEEE, 2018.

\end{thebibliography}
}

\end{document}